\documentclass{article}
\usepackage{spconf}
\usepackage{amsmath,graphicx}
\usepackage{amsfonts}
\usepackage{bbm}
\usepackage{booktabs}
\usepackage{amssymb,amsmath,bm}
\usepackage{textcomp}
\usepackage{algorithm}
\usepackage{algpseudocode}
\usepackage{float}
\usepackage{multirow}
\usepackage{mathtools}
\usepackage{placeins}
\usepackage{subcaption}  % A more recent package
\usepackage{caption}

\graphicspath{../figures}

\def\blanksym{{\ensuremath{\langle b \rangle}}}

\newcommand{\ssf}[1]{%
  \bm{\mathsf{#1}}%
}

\usepackage{tikz,placeins}
\usetikzlibrary{arrows,decorations.markings}
\tikzstyle{every picture}+=[font=\rmfamily\it\bfseries\large]
\usetikzlibrary{positioning, shapes}
\pgfdeclarelayer{background}
\pgfdeclarelayer{foreground}
\pgfsetlayers{background,main,foreground}

\newcommand{\specialcell}[2][c]{%
     \begin{tabular}[#1]{@{}c@{}}#2\end{tabular}}
\newcommand{\specialcelll}[2][c]{%
     \begin{tabular}[#1]{@{}l@{}}#2\end{tabular}}

\DeclareUnicodeCharacter{2212}{-}

\title{ A review of on-device fully neural end-to-end automatic speech
recognition algorithms}
\name{Chanwoo Kim, Dhananjaya Gowda, Dongsoo Lee, Jiyeon Kim, 
  \thanks{Thanks to Samsung Electronics for funding this research. 
The authors are thankful to President Sebasitan Seung, Executive Vice President Seunghwan Cho 
and speech processing Lab. members at Samsung Research.}}
\secondlinename{Ankur Kumar, Sungsoo Kim, Abhinav Garg, and Changwoo Han}%Jiyeon Kim, Abhinav Garg, Sungsoo Kim, and Changwoo Han}

\address{Samsung Research, Seoul, South Korea \\
  {\small \tt \{chanw.com, d.gowda, dongsoo3.lee, jstacey7.kim, } \\
  {\small \tt  ankur.k, ss216.kim, abhinav.garg, cw1105.han\}@samsung.com }}
\begin{document}
\ninept
\maketitle
\begin{abstract}
  In this paper, we review various end-to-end automatic speech recognition
  algorithms and their optimization techniques for on-device applications.
  Conventional speech recognition systems comprise a large number of discrete
  components such as an acoustic model, a language model, a pronunciation model, 
  a text-normalizer, an inverse-text normalizer, a decoder based on a Weighted Finite State
  Transducer (WFST), and so on. To obtain sufficiently high speech recognition
  accuracy  with such conventional speech recognition systems, a very large
  language model (up to 100 GB) is usually needed. Hence, the corresponding
  WFST size becomes enormous, which prohibits their on-device implementation. Recently,
fully neural network end-to-end speech recognition algorithms have been
  proposed. Examples include speech recognition systems based on  Connectionist Temporal Classification (CTC), Recurrent Neural Network Transducer (RNN-T), Attention-based Encoder-Decoder models (AED), Monotonic
  Chunk-wise Attention (MoChA), 
  transformer-based speech recognition systems, and so on. These fully neural
  network-based systems require much smaller memory footprints compared to
  conventional algorithms, therefore their on-device implementation has become
  feasible. In this paper, we review such end-to-end speech recognition models.
  We extensively discuss their structures, performance, and advantages compared
  to conventional algorithms.
\end{abstract}
\section{Introduction}
\label{sec:intro}
The advent of deep learning techniques has dramatically improved accuracy of
speech recognition models \cite{G_Hinton_IEEE_Signal_Process_Mag_2012}.
Deep learning techniques first saw success by replacing
the Gaussian Mixture Model (GMM) of the Acoustic Model (AM) part of the
conventional speech recognition systems
\cite{l_r_rabiner_proceedings_of_ieee_1989_00} with the Feed-Forward Deep Neural Networks 
(FF-DNNs), further with Recurrent Neural Network (RNN) such as the 
Long Short-Term Memory (LSTM) networks
\cite{s_hochreiter_neural_computation_1997_00} or Convonlutional Neural
Networks (CNNs). In addition to this, there have been improvements in noise
robustness by using models motivated by auditory processing
\cite{c_kim_taslp_2016_00, c_kim_interspeech_2014_00,
C_Kim_INTERSPEECH_2009_2}, data augmentation techniques
\cite{C_Kim_ASRU_2009_2, d_park_interspeech_2019_00, c_kim_interspeech_2018_00}, and beam-forming \cite{j_heymann_icassp_2016_00}.
Thanks to these advances, voice assistant devices such as Google Home
\cite{c_kim_interspeech_2017_00} and Amazon Alexa
have been widely used at home environments.

Nevertheless, it was not easy to run such high-performance speech recognition systems 
on devices largely because of the size of the Weighted Finite State Transducer (WFST) 
handling the lexicon and the language model. 
Fortunately, all-neural end-to-end (E2E) speech recognition systems
were introduced which do not need a large WFST or an n-gram Language Model (LM) 
\cite{j_chorowski_nips_2015_00}.
These complete end-to-end systems have started surpassing the performance of
the conventional WFST-based decoders with a very large training 
dataset \cite{c_chiu_icassp_2018_00} and a better choice of target unit 
such as Byte Pair Encoded (BPE) subword units.

In this paper, we provide a comprehensive review of the various components and algorithms of an end-to-end speech recognition system.
In Sec.~\ref{sec:nn_components}, we give a brief overview of the various neural building blocks of an E2E Automatic Speech Recognition (ASR) model.
The most popular E2E ASR architectures are reviewed in Sec.~\ref{sec:architectures}.
Additional techniques used to improve the performance of E2E ASR models are discussed in Sec.~\ref{sec:enhance}.
Techniques used for compression and quantization of the all-neural E2E ASR models are covered in Sec.~\ref{sec:compress}.
Sec.~\ref{sec:conclude} gives a summary of the paper.
%
%
%# Data augmentation and overfitting
\section{Neural network components for end-to-end speech recognition}
\label{sec:nn_components}
\begin{figure*}
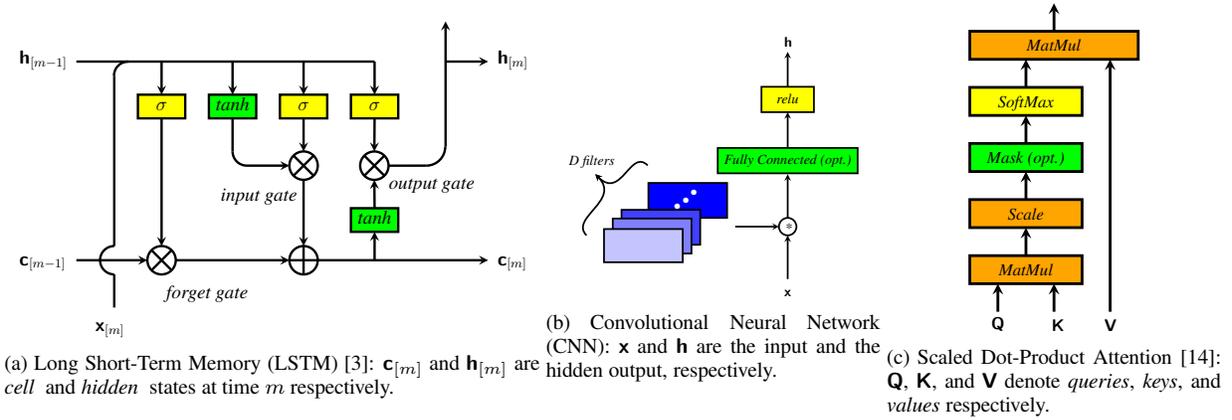

  \centering
  \begin{subfigure}{0.4\textwidth}
    \centering
    \resizebox{70mm}{!}{
      \input{lstm_structure.tex}
    }
    \caption {
      Long Short-Term Memory (LSTM)
      \cite{s_hochreiter_neural_computation_1997_00}:
      $\ssf{c}_{[m]}$ and $\ssf{h}_{[m]}$ are {\it cell } and 
      {\it hidden }
      states at time $m$
      respectively.
      \label{fig:lstm_structure}
    }
  \end{subfigure}
  \begin{subfigure}{0.25\textwidth}
    \centering
    \resizebox{40mm}{!}{
      \input{cnn_structure.tex}
    }
    \caption {
      Convolutional Neural Network (CNN): $\ssf{x}$ and $\ssf{h}$ are the input and the hidden output, respectively.
      \label{fig:cnn_structure}
    }
  \end{subfigure}
  \begin{subfigure}{0.25\textwidth}
    \centering
    \resizebox{25mm}{!}{
      \input{self_attention_structure.tex}
    }
    \caption {
      Scaled Dot-Product Attention \cite{a_vaswani_nips_2017_00}:
      $\ssf{Q}$, $\ssf{K}$, and $\ssf{V}$ denote {\it queries}, {\it keys}, and  {\it values} 
      respectively.
      \label{fig:self_attention_structure}
    }
  \end{subfigure}
  \captionsetup{justification=centering}
  \caption{
    Structures of neural net components frequently used for end-to-end speech
    recognition: \\ (a) LSTM, (b) CNN, and (c) Scaled Dot-Product Attention.
  }
  \label{fig:components}
  \vspace{-5mm}
\end{figure*}
\begin{table}[t]
\centering
\caption{Size of the intermediate buffer required for streaming speech recognition \cite{j_park_slt_2021_00}. $L$ is the sequence length, $D$ is the unit size or the dimension of each layer, and $T$ is the filter width of convolution. Refer to Sec. \ref{sec:nn_components} about how to obtain the {\it typical values} shown in this table. }
  \label{tbl:nn_components}
\centering
\begin{tabular}{l c c c}
  \toprule
  \multicolumn{1}{c}{Model }
                                      & \multicolumn{1}{c}{
                                        \specialcell{Memory \\ Footprint}}
                                      &  \multicolumn{1}{c}{Typical Value}
                                      & \multicolumn{1}{c}{
                                          \specialcell{Intra-Sequence \\ Parallelism}}
                                       \\
  \midrule
  LSTM   & $O(ND)$ & 98KB  & X\\
  Convolution &  $O(TND)$ & 245KB & O\\
  Self-attention & $O(LND)$ & 12MB & O\\
  \bottomrule
  \vspace{-5mm}
\end{tabular}
\end{table}
%
%
%In this section, we give an overview of the neural network components
%employed in all-neural end-to-end speech recognition systems. 
Fig.  \ref{fig:components} illustrates neural network components commonly 
employed in end-to-end speech recognition systems. Gated RNNs such as LSTMs 
and Gated Recurrent Units (GRUs) \cite{k_cho_emnlp_2014_00}  have been used 
from the early days of encoder-decoder \cite{i_sutskever_nips_2014_00} and
attention-based models. The operation of an LSTM
is described by the following equation:

\newcommand{\prevhx}[1]{{\ensuremath{\ssf{W}_#1  \ssf{x}_{[m]}
  + \ssf{U}_#1 \ssf{h}_{[m-1]}
+ \ssf{b}_#1}}}

\begin{subequations}
  \begin{align}
    \ssf{i}_{[m]} & =  \sigma \big(\prevhx{i} \big) \\
    \ssf{f}_{[m]} & =  \sigma \left(\prevhx{f} \right) \\
    \ssf{o}_{[m]} & =  \sigma \left(\prevhx{o} \right) \\
    \tilde{\ssf{c}}_{[m]} & = \tanh \left(\prevhx{c}  \right) \\
    \ssf{c}_{[m]} & = \ssf{i}_{[m]} \odot \tilde{\ssf{c}}_{[m]}
    + \ssf{f}_{[m]} \odot \ssf{c}_{[m-1]} \\
    \ssf{h}_{[m]} & = \ssf{o}_{[m]} \odot \tanh(\ssf{c}_{[m]}),
  \end{align}
  \label{eq:lstm}
\end{subequations}
where $\ssf{W}_{(\cdot)}$ and $\ssf{U}_{(\cdot)}$ are weight matrices,
$\ssf{b}_{(\cdot)}$ is the bias vector. $\ssf{i}[m]$, $\ssf{f}[m]$, and
$\ssf{o}_{[m]}$ are the input, forget, and output gates at time $m$, respectively.
$\sigma(\cdot)$ is the sigmoid function and $\odot$ is the Hadamard product
operator. $\ssf{c}_{[m]}$ and $\ssf{h}_{[m]}$ are the cell and hidden states. Fig.
\ref{fig:lstm_structure} shows the structure of an LSTM described by
\eqref{eq:lstm}.
One of notable 
advantages of gated RNNs is that they require relatively smaller memory 
footprint compared to other models as shown
in Table \ref{tbl:nn_components}. Their another
advantage is streaming capability if unidirectional models (\emph{e.g.}
uni-directional LSTMs or GRUs) are employed. Because of these advantages, 
at the time of writing this paper, most of the commercially available end-to-end on-device speech recognition systems
are based on LSTMs  \cite{k_kim_asru_2019_00, y_he_icassp_2019_00}.  
However, as shown in Table \ref{tbl:nn_components}, LSTMs have disadvantages 
in terms of {\it intra-sequence parallelism}, since the computation needs to be
done sequentially.

Various CNN-based approaches have been successfully employed in building
end-to-end speech recognition systems \cite{j_park_nips_2018_00,
a_hannun_interspeech_2019_00}. These approaches are characterized by a group
of filters and a nonlinear rectifier as shown in Fig. \ref{fig:cnn_structure}.
 As an example of various CNN-based approaches, depthwise 1-dimensional CNN
 is represented by the following equations \cite{j_park_slt_2021_00}:
\begin{subequations}
  \begin{align}
    x'_{[m],\,d} & =  \sum_{t=-(T-1)/2}^{(T-1)/2}  {W_{t,d}} \cdot {x_{[m+t],\, d}},\quad
    0 \le d \le D- 1\\
    \ssf{x}'_{[m]} & = \big[ x'_{[m],\, 0},  x'_{[m],\,1},\, \cdots \,,
    x'_{[m],\,{D-1}}\big]^{\intercal} \\
    {\ssf{h}_{[m]}} & = \text{relu}({\ssf{V}\ssf{x}'_{[m]}+ \ssf{b}}),
    \label{eq:rect}
  \end{align}
  \label{eq:conv}
\end{subequations}
where $T$ and $D$ are the length of the 1-dimensional filter and the number of
such filters respectively. $x_{[m],\,d},\,0 \le d \le D-1$ is the $d$-th element of
the input $\ssf{x}_{[m]}$ at the time index $m$.
$\ssf{W} \in \mathbb{R}^{T \times D}$, 
$\ssf{V} \in \mathbb{R}^{D \times D}$, and  $\ssf{b} \in \mathbb{R}^{D}$
are trainable variables.  Unlike gated RNNs, to calculate the output of the current time step, CNN does not require the completion of computation for the previous time steps,
 which enables {\it intra-sequence parallelism} as summarized in Table
 \ref{tbl:nn_components}. Thus, CNN-based
end-to-end speech recognition systems have advantages in computational
efficiency on embedded processors supporting Single Instruction Multiple Data
(SIMD) by exploiting {\it intra-sequence parallelism} \cite{j_park_slt_2021_00}.

Recently, {\it self-attention} has been also successfully applied to speech recognition.
In \cite{a_vaswani_nips_2017_00}, self-attention mechanism was implemented
using a scaled-dot attention described by the following equation:
\begin{align}
  \text{Attention}(\ssf{Q}, \ssf{K}, \ssf{V})
  = \text{softmax}\left(\frac{\ssf{Q}\ssf{K}^{\intercal}}{\sqrt{d_k}}\right)
  \ssf{V},
  \label{eq:attention_eq}
\end{align}
where $\ssf{Q}$, $\ssf{K}$, and $\ssf{V}$ are matrices representing a {\it query}, a {\it key},
and a {\it value}, and $d_k$ is the dimension of the {\it key}. In
a self-attention layer, all of the {\it queries}, {\it keys}, and {\it values}
are the outputs of the previous layer.

In Table \ref{tbl:nn_components}, we compare typical values of memory 
footprint required for stacks of these neural network layers
\cite{j_park_slt_2021_00}. These stacks correspond 
to the neural net layers in Fig. \ref{fig:ctc} or the encoder portion of 
Recurrent Neural Network-Transduer (RNN-T) \cite{a_graves_corr_2012_00, a_graves_icassp_2013_00} or {\it attention-based} models in Fig. \ref{fig:rnn_t_diagram} and Fig. \ref{fig:attention_diagram}, respectively. 
In obtaining these values, we assume that the number
of layer $N$ is 6 for the LSTM case \cite{k_kim_asru_2019_00}, and 
15 for the CNN and the self-attention cases. 
For speech recognition, $L$ is usually a few hundred, while $T$ is
about ten or less.  Based on this, we assume that the dimension $D$ is 2048, 
the sequence length $L$ is 100, and the filter length $T$ is 5, respectively, 
to calculate typical values \cite{j_park_slt_2021_00}.
\section{End-to-End speech recognition architectures}
\label{sec:architectures}
\begin{figure*}
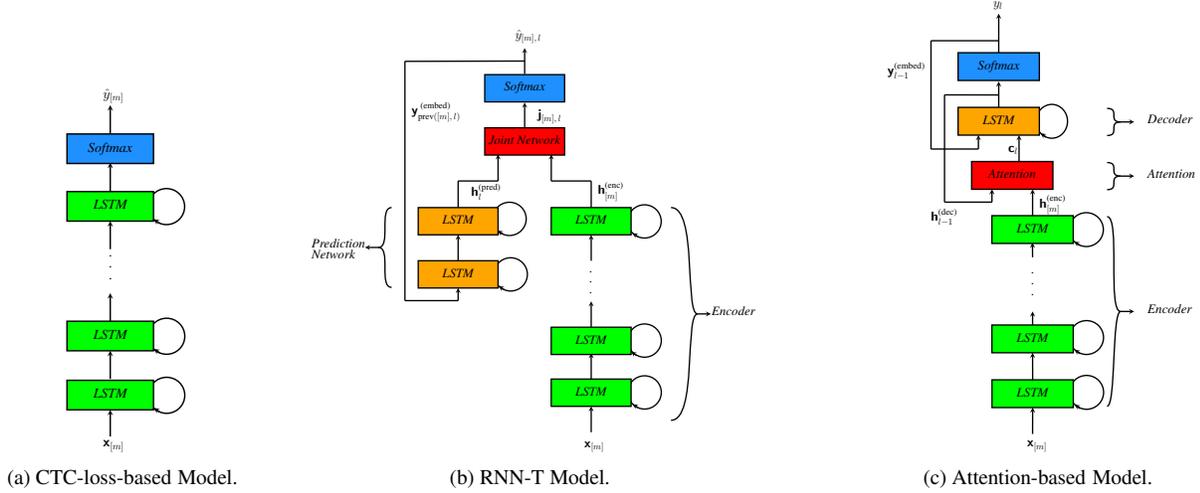

  \centering
  \begin{subfigure}[b]{0.20\textwidth}
    \centering
    \resizebox{17mm}{!}{
      \input{ctc_diagram.tex}
    }
    \caption {
      CTC-loss-based Model.
      \label{fig:ctc}
    }
  \end{subfigure}
  \begin{subfigure}[b]{0.40\textwidth}
    \centering
    \resizebox{60mm}{!}{
      \input{rnn_t_diagram.tex}
    }
    \caption {
      RNN-T Model.
      \label{fig:rnn_t_diagram}
    }
  \end{subfigure}
  \begin{subfigure}[b]{0.35\textwidth}
    \centering
    \resizebox{42mm}{!}{
      \input{attention_diagram.tex}
    }
    \caption {
      Attention-based Model.
      \label{fig:attention_diagram}
    }
  \end{subfigure}
  \caption{
    Comparison of block diagrams of different sequence-to-sequence speech
    recognition approaches.
  }
  \vspace{-5mm}
  \label{fig:architectures}
\end{figure*}
Speech recognition is a task of finding the {\it sequence-to-sequence} mapping
from an input sequence of acoustic features to a output sequence of labels. Let
us denote the input and output sequences by $ \ssf{x}_{[0:M]}$ and $y_{0:L}$ 
as shown below:
\begin{subequations}
  \begin{align}
    \ssf{x}_{[0:M]} & = \left[\ssf{x}_{[0]},\, \ssf{x}_{[1]},\,
    \ssf{x}_{[2]},\, \cdots, \,
    \ssf{x}_{[M-1]}\right], \\
    y_{0:L} & = \big[y_0,\,y_1,\,y_2,\, \cdots,\,y_{L-1}\big],
  \end{align}
  \label{eq:in_out}
\end{subequations}
where $M$ and $L$ are the lengths of the input acoustic feature sequence and 
the output label sequence, respectively. The sequence notation adopted in this
paper including \eqref{eq:in_out}
follows the {\tt Python} array slice notation. In this paper, by convention, we use
a {\it  bracket} to represent a periodically sampled sequence such as the acoustic feature, 
and use a {\it subscript} to represent a non-periodic sequence such as the output label.
Fig. \ref{fig:architectures} shows structures of end-to-end all neural
speech recognition systems. Even though we use a stack of LSTMs in Fig. 
\ref{fig:architectures}, any kinds of neural network components described in
Sec. \ref{sec:nn_components} may be employed instead of LSTMs. 
\subsection{Connectionist Temporal Classification (CTC)}
\label{sec:ctc}
The simplest way of implementing an 
end-to-end speech recognizer is using a stack of neural network layers with
a Connectionist Temporal Classification loss \cite{a_graves_icml_2006_00} as
shown in Fig.  \ref{fig:ctc}. This model defines a probability distribution
over the set of output labels augmented with a special blank symbol, $\blanksym$.
We define the set of all possible alignments 
$\mathcal{B}^{\text{CTC}}\left(\ssf{x}_{[0:M]},\, y_{0:L}\right)$, as the set of all label
sequences $\ssf{z}_{[0:M]}$ where  $z[m] \in \mathcal{Y} \cup 
{\langle b \rangle},\,0\le m < M$, such that
$\ssf{z}_{[0:M]}$ is identical to $y_{0:l}$ after removing all blank symbols
$\blanksym$. $\mathcal{Y}$ is the set of the entire alphabet of the output
labels. Under this assumption, the posterior probability of the output label sequence is given by the following equation: 
\begin{align}
  P\left( y_{0:L} \big| \ssf{x}_{[0:M]} \right) = \sum_{
    \substack{z_{[0:M]} \in \\ \mathcal{B}^{\text{CTC}}\left(\ssf{x}_{[0:M]},\,
    y_{0:L}\right) }}
  { \Pi_{m=0}^{M-1} P\left(z_{[m]} \big| \ssf{x}_{[0:M]} \right)  }.
      \label{eq:ctc_prob}
\end{align}
The CTC loss is defined by the following equation:
\begin{align}
  \mathbbm{L}_{\text{CTC}} = - E 
    \Big[
      \log 
        \left( 
          P\left( y_{0:L} \big| \ssf{x}_{[0:M]} \right) 
        \right) 
    \Big].
\end{align}
The parameters of a model with the CTC loss are updated using the {\it
forward-backward recursion} assuming conditional indepedence
\cite{a_graves_icml_2006_00}, which is similar to the 
forward-backward algorithm used for training Hidden Markov Models (HMMs) 
\cite{l_r_rabiner_proceedings_of_ieee_1989_00}.

\subsection{Recurrent Neural Network-Transducer (RNN-T)}
\label{sec:rnn_t}
In spite of its simplicity, the model described in Sec. \ref{sec:ctc} has several shortcomings including the conditional independence assumption 
during the model training \cite{s_kim_icassp_2017_00} and the lack of explicit feedback paths from the output label.
An improved version of this model is the RNN-T model \cite{a_graves_icassp_2013_00} 
shown in Fig.  \ref{fig:rnn_t_diagram}. In this model, 
there is an explicit feedback path
from the output label to the prediction network which plays a similar role to
an LM. The probability model of an RNN-T based model is similar to the CTC
model:
\begin{align}
  P\left( y_{0:L} \big| \ssf{x}_{[0:M]} \right) = \sum_{
    \substack{z_{[0:M+L]} \in \\ \mathcal{B}^{\text{RNN-T}}\big(\ssf{x}_{[0:M]},\,
    y_{0:L}\big) }}
  { \Pi_{u=0}^{M+L-1} P\left(z_{[u]} \big| \ssf{x}_{[0:M]} \right)  },
      \label{eq:ctc_prob}
\end{align}
where $\mathcal{B}^{\text{RNN-T}}\left(\ssf{x}_{[0:M]},\, y_{0:L}\right)$ is the set of all possible label sequences $\ssf{z}_{[0:M+L]}$ where  $z[u] \in \mathcal{Y} \cup 
{\langle b \rangle},\, 0\le u < M+L$, such that after removing blank symbols $\langle
b \rangle$ from $\ssf{z}_{[0:M+L]}$, it becomes the same as $y_{0:L}$. As in
Sec. \ref{sec:ctc}, $\mathcal{Y}$ is the set of the entire alphabet of the
output labels.

\subsection{Attention-based Models}
\label{sec:attention}
Another popular end-to-end speech recognition approach is employing the attention-mechanism
as shown in Fig. \ref{fig:attention_diagram} \cite{j_chorowski_nips_2015_00}. In attention-based approaches, we use  
the attention between the encoder and decoder hidden outputs. The equation for
encoder-decoder attention is basically the same as \eqref{eq:attention_eq}:
\begin{subequations}
  \begin{align}
    e_{[m],\, l} & = \text{\it Energy}(\ssf{h}^{(\text{\it enc})}_{[m]},\ssf{h}^{(\text{\it
    dec})}_{l-1}) \label{eq:att_energy} \\
    a_{[m],\, l} & = \text{softmax}(e_{[m],\, l}) \\
    \ssf{c}_l & = \sum_{m=0}^{M-1} a_{[m],\, {l}} \ssf{h}^{(\text{\it
    enc})}_{[m]},
  \end{align}
\end{subequations}
where $\ssf{c}_l$ is the {\it context vector} which is used as the input to the decoder.
$\ssf{h}^{\text{({\it enc})}}[m]$ and $\ssf{h}^{\text{({\it dec})}}_{l-1}$
are hidden outputs from the encoder at time $m$ and from the decoder at the label
index $l-1$, respectively.
$e_{[m],\, l}$  in  \eqref{eq:att_energy} is the {\it energy} for the input
time index $m$ and the output label index $l$.
\subsection{Monotonic Chunkwise Attention(MoChA)-based Models}
\label{sec:mocha}
\begin{figure}
  \centering
    \resizebox{70mm}{!}{
      \input{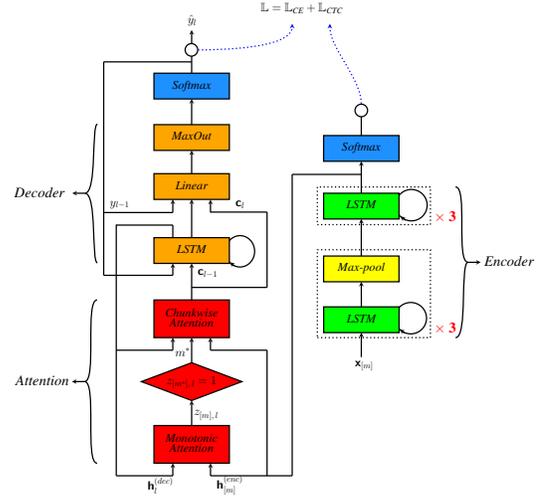}
  }
    \captionsetup{justification=centering}
  \caption{
    The structure of an on-device speech recognition system based on
    Monotonic Chunkwise Attention (MoChA) \cite {k_kim_asru_2019_00}
  }
  \vspace{-5mm}
  \label{fig:asru_2019_attention_diagram}
\end{figure}

Although the attention-based approach in \ref{sec:attention} has been quite
successful, on-line streaming recognition with this approach has been 
a challenge. This is because the entire sequence of input features 
must be encoded before the decoder starts generating the first output
label. Several variations of the attention including Monotonic Chunkwise
Attention (MoChA) \cite{c_chiu_iclr_2018_00} 
have been proposed to resolve this problem.
In the MoChA model, there
are two attention mechanisms: a hard monotonic attention followed by a soft chunkwise
attention. The hard monotonic attention is employed to determine which element
should be attended from a sequence of hidden encoder outputs
$\ssf{h}^{\text{({\it enc})}}_{[m]}$.
The hard monotonic attention is obtained from the hidden encoder output
$\ssf{h}^{\text{({\it enc})}}_{[m]}$ at the time index $m$ and the hidden
decoder output $\ssf{h}^{(\text{\it dec})}_{l-1}$ at the output label index
$l-1$ as follows:
\begin{subequations}
  \begin{align}
    e^{\text{({\it mono})}}_{[m],\,l} & = \text{\it MonotonicEnergy}
      (\ssf{h}^{(\text{\it enc})}_{[m]}, \ssf{h}^{(\text{\it dec})}_{l-1}) \\
    a^{\text{({\it mono})}}_{[m], l} & = \sigma(e^{\text{(\it mono)}}_{[m],\,
    {l}}) \\
    z_{[m], l} & \sim \text{\it Bernoullli}
      \left(a^{(\text{\it mono})}_{[m],\,{l}}\right),
    \label{eq:hard_output}
  \end{align}
\end{subequations}
where $\sigma(\cdot)$ is a logistic sigmoid function and {\it MonotonicEnergy} is the
energy function defined as follows \cite{c_chiu_iclr_2018_00}:
\begin{align}
  \text{\it MonotonicEnergy}&(\ssf{h}^ {\text{({\it enc})}}_{[m]},\;\ssf{h}^{\text{({\it dec})}}_{l-1})
  \nonumber \\
  \;  = g \frac{v^{\intercal}}{\| v \|} & \tanh (W^{\text{({\it dec})}}
  \ssf{h}^{\text{({\it dec})}}_{l-1} + W^{\text{({\it enc})}}  \ssf{h}^
  {\text{({\it enc})}}_{[m]} + b ) + r,
\end{align}
where $v, W^{\text{({\it dec})}}, W^{\text{({\it enc})}}, b, g$, and $r$ are learnable variables. After finding out the
position to attend using \eqref{eq:hard_output}, a soft attention with a fixed
chunk size is employed to generate the context vector that will be given as an
input to the decoder.
We refer readers interested in the more detailed structure of MoChA to 
\cite{k_kim_asru_2019_00, c_chiu_iclr_2018_00}.
A block schematic of an on-device speech recognizer based on MoChA is shown in Fig.~\ref{fig:asru_2019_attention_diagram}.
\subsection{Comparison between MoChA and RNN-T based models}
\label{sec:comp_mocha_rnn_t}
In this section, we compare a MoChA-based model described in Sec.
\ref{sec:mocha} with a RNN-T-based model discussed in Sec. \ref{sec:rnn_t}.
For these RNN-T and MoChA models, we used the same encoder structures 
consisting of six LSTM layers with the unit size of 1,024. 
The lower three LSTM layers in the encoder are interleaved with 2:1
{\it max-pool} layers as in \cite{k_kim_asru_2019_00}.
A single LSTM layer with the unit size of 1024 is employed for both the decoder
of the MoChA-based model and the prediction network of RNN-T-based model. 
Training was performed using an {\it in-house} training
toolkit built with the same {\tt Keras} \cite{f_chollet_keras_2015_00} and {\tt
Tensorflow 2.3} APIs \cite{m_abadi_usenix_2016}. 
A 40-dimension {\it power-mel} feature with a power coefficient of 1/15
\cite{c_kim_taslp_2016_00} is used as the input feature. We prefer the {\it power-mel}
feature to the more frequently used {\it log-mel} feature since
the power-mel feature shows better speech recognition accuracy
\cite{c_kim_interspeech_2019_00, c_kim_asru_2019_00, c_kim_asru_2019_01}.
In Table \ref{tbl:mocha_rnnt_perf_comparison},  we compare the performance of 
MoChA and RNN-T in terms of speech recognition accuracy and latency.
The model is compressed by 8-bit quantization and Low Rank Approximation
(LRA) \cite{d_lee_arxiv_2018_00}. More details about the compression procedure
is described in \cite{k_kim_asru_2019_00}. The latency was measured on a 
{\tt Samsung Galaxy Note 10} device.
As shown in this table, the MoChA-based model shows slightly better
speech recognition accuracy compared to the RNN-T-based model. 
However, the RNN-T-based model is noticeably better than the MoChA-based
model in terms of latency and consistency of latency. The MoChA-based model
shows more variation in latency compared to the RNN-T based model.
\begin{table}[!htbp]
  \renewcommand{\arraystretch}{1.3}
  \centering
  \caption{\label{tbl:mocha_rnnt_perf_comparison}
  Performance comparison between the MoChA-based model and the RNN-T based model on Librispeech \texttt{test-clean} evaluation set.
  }
  \begin{tabular}{ l  c   c }
      \toprule
                              Model 
                             %& \specialcell{LibriSpeech\\ \footnotesize \texttt{test-clean}\\WER                             (\%)}
                             & \specialcell{WER                           }
                             & \specialcell{ Avg. Latency } \\
      \midrule
               MoChA  &   6.88  \% &    225 ms  \\
               RNN-T  &   7.63  \% &    86.5 ms  \\
      \bottomrule
  \end{tabular}
  \vspace{-2mm}
\end{table}
%
%
%
%
% Adds two-pass decoder diagram
\section{Approaches for further performance improvement}
\label{sec:enhance}
Recently, various techniques have been proposed to further improve the
performance of all neural end-to-end speech recognition systems described in
Sec.  \ref{sec:architectures}. We will discuss some of these techniques in this
section.
\subsection{Combination with a non-streaming model}
\begin{figure}
  \centering
    \resizebox{60mm}{!}{
      \input{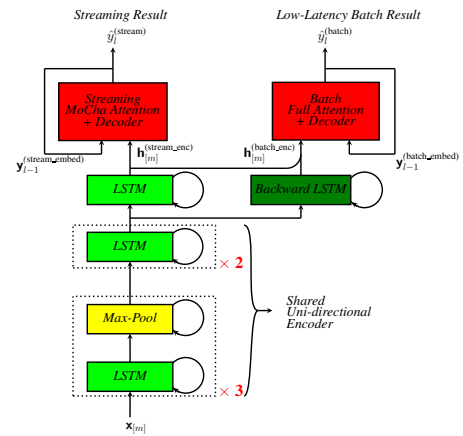}
  }
    \captionsetup{justification=centering}
  \caption{
    Speech recognition system that performs streaming recognition followed 
    by low-latency batch recognition \\ for improved performance. 
  }
  \label{fig:shallow_bfa_diagram}
  \vspace{-5mm}
\end{figure}
Even though Bidirectional LSTMs (BLSTMs)  perform significantly better than
unidirectional counterparts for on-device
applications \cite{k_kim_asru_2019_00}, latency requirement usually prohibits their usage. 
To get the advantage of BLSTMs without significantly affecting overall latency, 
we proposed a recognition system combining the streaming MoChA model with a batch 
full attention model in \cite{d_gowda_interspeech_2020_00}. The entire structure
of this system is shown in Fig. \ref{fig:shallow_bfa_diagram}. In this model, there is 
a shared streaming encoder consisting of uni-directional LSTMs. On one side,
the streaming MoChA attention and decoder can generate streaming speech
recognition result. On the other side, one backward LSTM layer on top of the
shared unidirectional layer comprise a BLSTM layer, which is followed by a full
attention model. When the user finishes speaking, then this full attention
model generates low-latency batch speech recognition result. The latency impact is relatively small since there is only one layer of backward LSTM layer in
the encoder. The experimental result is summarized in Table
\ref{tbl:shallow_bfa}. We use the same model architecture and
experiment configuration as described in Sec. \ref{sec:comp_mocha_rnn_t}.
As shown in this table, the proposed model in the bottom row shows
significantly better result than the MoChA-based model shown in the top row of
this table with endurable increase in latency.

\begin{table}[!htbp]
  \renewcommand{\arraystretch}{1.3}
  \centering
  \caption{\label{tbl:shallow_bfa}
  Word Error Rate (WER) comparison between a streaming model, a non-streaming model, and
  a streaming model combined with a single layer of backward LSTM followed by a full
  attention to minimize the impact on latency.
  }
  \begin{tabular}{ l  c  c }
      \toprule
                             Model 
                             & \specialcell{LibriSpeech\\ \footnotesize  \texttt{test-clean}}
                             & \specialcell{LibriSpeech\\ \footnotesize  \texttt{test-other}}
                             \\
      \midrule
               \specialcelll{\footnotesize Uni-Directional Encoder \\ \footnotesize \quad
               + MoChA-Attention}  &   6.88  \% &  19.11 \%   \\
               \specialcelll{\footnotesize Bi-Directional Encoder \\ \footnotesize \quad + Full-Attention}
               &   3.62  \% &  11.20 \%  \\
               \specialcelll{\footnotesize Shared Uni-Directional Encoder\\
               \footnotesize \quad
               + Full-Attention}  &   4.09  \% &  12.04  \%  \\
      \bottomrule
  \end{tabular}
  \vspace{-5mm}
\end{table}
\subsection{Shallow-fusion with language models}
\label{sec:lm}

End-to-end speech recognition models discussed in Sec.  \ref{sec:architectures}
are trained using only paired speech-text data. Compared to traditional 
AM-LM approaches where an LM is often trained using a much larger text corpus 
possibly containing billions of sentences \cite{a_kannan_icassp_2018_00}, 
the end-to-end speech recognition model sees much limited number of word sequences during 
the training phase. To get further performance improvement, researchers have
proposed various techniques of incorporating external language models such as
{\it shallow-fusion} \cite{j_chorowski_interspeech_2017_00}, {\it deep-fusion}
\cite{c_gulcehre_corr_2015_00}, and {\it cold-fusion}
\cite{a_sriram_corr_2017_00}. Among them, in spite of its simplicity, shallow-fusion
seems to be more effective than other approaches
\cite{s_toshniwal_slt_2018_00}. In shallow-fusion, the log probability from the
end-to-end speech recognition model is linearly interpolated with the
probability from the language model as follows:
\begin{align}
  \log p_{\text{sf}}\left( y_l \big| \ssf{x}_{[0:m]}  \right) & =  
    \log p \left(y_l \big| \ssf{x}_{[0:m]}, \hat{y}_{0:l} \right) \nonumber  + \lambda  \log p_{\text{lm}} \left( y_l \big|  \hat{y}_{0:l}\right)
\end{align}
where $p_{\text{lm}} \left( y_l \big|  \hat{y}_{0:l}\right)$ is the
probability of predicting the label $y_l$ from the LM, and the 
$p \left(y_l \big| \ssf{x}_{[0:m]}\right)$ is the posterior probability
obtained from the end-to-end speech recognition model.

Shallow-fusion with an LM is helpful in enhancing speech performance in general
domains, thus, most of the state of the art speech recognition results on
publicly available test sets ({\emph (e.g) LibriSpeech\cite{v_panayotov_icassp_2015_00}}) are obtained with this technique
\cite{d_park_interspeech_2019_00, c_kim_interspeech_2019_00, a_gulati_interspeech_2020_00}. In addition to this,
this technique is also very useful for enhancing performance in special domain
like personal names, geographic names, or music names by employing
domain-specific LMs. In case of n-gram LMs, they can be easily built on-device
to support personalized speech recognition \cite{k_kim_asru_2019_00}.
For on-device command recognition, shallow-fusion with a WFST is also useful for
specific domains since WFST contains a list of words not just subword units
\cite{a_garg_interspeech_2020_01}.
% Adds two-pass decoder diagram

\subsection{Improving NER performance using a spell corrector}
\label{sec:ner}

Even though end-to-end all neural speech recognition systems have shown quite
remarkable speech recognition accuracy with small memory footprint, it has been
frequently observed that performance is poorer in recognizing named entities
\cite{r_prabhavalkar_interspeech_2017_00}. Compared to conventional speech
recognizers built with a WFST containing dictionary information,
an end-to-end speech recognition system does not explicitly contain a list of
named entities. Therefore, the speech recognition accuracy of such systems
is generally low for specialized domains handling song names, 
composers' names,  personal names, and geographical names. Without 
dictionary information, spelling errors may also occur with all
neural speech recognizers. This named entity recognition issue can be somewhat relieved by applying
shallow-fusion with an LM as described in Sec. \ref{sec:lm}. However, more drastic
performance improvement is usually obtained by classifying the domain from the
speech recognition output and applying spell correction using a list of 
named entities found in that domain. In \cite{a_garg_interspeech_2020_01}, a multi-stage spell correction approach was proposed to handle a large list of named entities with on-device speech recognizers.

\section{Compression}
\label{sec:compress}
To run speech recognition models on embedded processors with limited memory, we often need to further reduce the parameter size in order to satisfy computational cost and memory footprint requirements. There are many known techniques to reduce the model size such as quantization, factorization, distillation, and pruning. The simplest way of accomplishing compression may be applying 8-bit quantization, which is supported by 
{\tt TensorflowLite$^{\text{TM}}$}. Several commercially available speech recognition systems \cite{k_kim_asru_2019_00, y_he_icassp_2019_00} have been built using this technique. In feed-forward neural networks, 2-bit or 3-bit quantization has been
successfully  employed \cite{k_hwang_sips_2014_00}. However, for general purpose processors, there are relatively small gains in going below 8-bit quantization, since Arithmetic Logic Units (ALUs) usually do not support sub 8-bit arithmetic. Another popular technique in reducing the model size of LSTMs is applying Low Rank Approximation (LRA) \cite{i_mcgraw_icassp_2016_00}. As an example, the {\it DeepTwist} algorithm, which is based on LRA, is employed to reduce the parameter size in our previous work \cite{k_kim_asru_2019_00, a_garg_interspeech_2020_01}. More specifically, we were able to reduce the size of MoChA-based models from 531 MB to less than 40 MB without sacrificing the performance using 8-bit quantization and LRA.
Pruning may be a good choice if hardware supports sparse matrix arithmetic. In \cite{s_han_iclr_2016_00}, authors propose a three-stage pipeline: 
pruning, trained quantization and Huffman coding, that work together 
to reduce the storage requirement of neural networks by 35x to 49x 
without affecting their accuracy. 
In \cite{r_pang_interspeech_2018}, the authors studied effectiveness of different compression techniques for end-to-end speech recognition models. They conclude that pruning is the most effective with proper hardware support. In the absence of this, for small models, distillation appears to be the best choice, while factorization appears to be the best approach for larger models.

\section{Conclusions}
\label{sec:conclude}
In this paper, we reviewed various end-to-end all neural automatic speech recognition
systems and their optimization techniques for on-device applications.
On-device speech recognition has huge advantages compared to the server-side
ones in terms of user privacy, operation without internet, server-cost, and latency. 
To operate speech recognition systems on embedded processors, we need to 
consider several factors such as recognition accuracy, computational cost, latency, 
and the model size.
We compared pros and cons of different neural network components such as Long
Short-Term Memory (LSTM), Convolutional Neural Network (CNN), 
 and attention mechanism.
We explained and compared different end-to-end neural speech recognition architectures such as 
a stack of LSTM layers with the Connectionist Temporal Classification (CTC)
loss \cite{a_graves_icml_2006_00}, 
Recurrent Neural Network-Transformer(RNN-T) \cite{a_graves_corr_2012_00,
a_graves_icassp_2013_00},
attention-based models, and models based on Monotonic Chunk-wise Attention
(MoChA) \cite{c_chiu_iclr_2018_00}. Further
improvement is achieved by combining a streaming model with a low-latency non-streaming model, 
by applying shallow-fusion with a Language Model (LM), and by applying spell correction using
a list of named entities \cite{a_garg_interspeech_2020_00}. We also discussed several model compression
techniques including quantization, singular value decomposition, pruning, and
knowledge distillation. 
These recent advances in all neural end-to-end speech recognition made it
possible to commercialize all neural on-device end-to-end speech recognition
systems  \cite{k_kim_asru_2019_00, y_he_icassp_2019_00, a_garg_interspeech_2020_01}.
%

% \cite{t_sainath_icassp_2020_00}
% \cite{s_han_iclr_2016_00}
% \cite{t_sainath_interspeech_2019_00}

% \cite{r_pang_interspeech_2018}

%
%
%
%\section{Conclusions}
%In this paper, we describe various end-to-end speech recognition models
%and discuss how they are employed for on-device applications. 
%For on-device speech recognition applications, streaming capability is 
%usually required to meet the latency requirement.
%We compare the performance of the MoCha-based model with the RNN-T based model
%in terms of speech recognition accuracy, inference time, latency, and model
%size.
%
%
%
% References should be produced using the bibtex program from suitable
% BiBTeX files (here: strings, refs, manuals). The IEEEbib.bst bibliography
% style file from IEEE produces unsorted bibliography list.
% -------------------------------------------------------------------------
%\ninept
\bibliographystyle{IEEEtran}
%\bibliography{../../common_bib_file/common_bib_file}
\bibliography{common_bib_file}

\end{document}